\documentclass[10pt,twocolumn,letterpaper]{article}

\usepackage{wacv}              

\usepackage{graphicx}
\usepackage{amsmath}
\usepackage{amssymb}
\usepackage{booktabs}

\usepackage{algorithm}
\usepackage{algpseudocode}
\usepackage{stackengine}
\usepackage{multirow}
\usepackage{tabularx}
\usepackage{subcaption}
\usepackage{esvect}
\usepackage{makecell}
\usepackage{overpic}
\usepackage{wrapfig}
\usepackage{tikz, xcolor}
\usetikzlibrary{calc,spy}

\usepackage[pagebackref,breaklinks,colorlinks]{hyperref}

\usepackage[capitalize]{cleveref}
\crefname{section}{Sec.}{Secs.}
\Crefname{section}{Section}{Sections}
\Crefname{table}{Table}{Tables}
\crefname{table}{Tab.}{Tabs.}

\tikzstyle{smallwindow} = [white, line width=0.30mm]
\tikzstyle{largewindow_w} = [white, line width=0.30mm]
\tikzstyle{largewindow_r} = [red, line width=0.30mm]

\tikzstyle{closeup_w} = [
  opacity=1.0,          
  height=1.2cm,         
  width=2.0cm,          
  connect spies, white  
]

\tikzstyle{closeup_r} = [
  opacity=1.0,          
  height=1.0cm,         
  width=1.4cm,          
  connect spies, white  
]

\def\wacvPaperID{49} 
\def\confName{WACV}
\def\confYear{2024}

\begin{document}

\title{Learning Residual Elastic Warps for  Image Stitching \\ under Dirichlet Boundary Condition}

\author{
Minsu Kim$^1$, \;\; Yongjun Lee$^2$, \;\; Woo Kyoung Han$^1$, \; and \; Kyong Hwan Jin$^2$\thanks{Corresponding author.} \\
$^1$DGIST, Republic of Korea \; $^2$Korea University, Republic of Korea \\
$^1${\tt\small \{axin.kim, cjss7894\}@dgist.ac.kr} \;
$^2${\tt\small \;\{lyj9805, kyong\_jin\}@korea.ac.kr} \\
}
\maketitle

\def\delequal{\mathrel{\stackon[1pt]{$=$}{$\scriptscriptstyle\Delta$}}}

\begin{abstract}
Trendy suggestions for learning-based elastic warps enable the deep image stitchings to align images exposed to large parallax errors. Despite the remarkable alignments, the methods struggle with occasional holes or discontinuity between overlapping and non-overlapping regions of a target image as the applied training strategy mostly focuses on overlap region alignment. As a result, they require additional modules such as seam finder and image inpainting for hiding discontinuity and filling holes, respectively. In this work, we suggest Recurrent Elastic Warps (REwarp) that address the problem with Dirichlet boundary condition and boost performances by residual learning for recurrent misalign correction. Specifically, REwarp predicts a homography and a Thin-plate Spline (TPS) under the boundary constraint for discontinuity and hole-free image stitching. Our experiments show the favorable aligns and the competitive computational costs of REwarp compared to the existing stitching methods. Our source code is available at \url{https://github.com/minshu-kim/REwarp}.
\end{abstract}

\section{Introduction}

Estimation of geometric transformations for image stitching requires descriptions of the spatial correspondences between given scenes. To calculate such relationships, existing methods are categorized into two branches: feature-based methods \cite{gao2011constructing, jia2021leveraging, joo2015line, li2017parallax, liao2019single, lin2011smoothly, zaragoza2013projective, liao2022natural} and learning-based \cite{kweon2021pixel, nie2020learning, nie2021unsupervised, nie2023learning, song2021end, song2022weakly, lai2019video} methods. Feature-based approaches detect key points or lines to match the textures of images in order to estimate the optimal transformation. In contrast, learning-based frameworks compute a 4D cost volume \cite{xu2017accurate} which computes receptive field-wise correlations and provides a prior like dense matching between given scenes.

The two branches have clearly distinct limitations. The recent feature-based frameworks \cite{jia2021leveraging, liao2019single} leverage line and key point matching and estimate transformation minimizing the error of the matched features keeping geometric structures. However, the approaches cause frequent failures and performance degradation in challenging environments like low resolution and few textured scenes.

\begin{figure}[t]
    \centering
    \includegraphics[width=\linewidth]{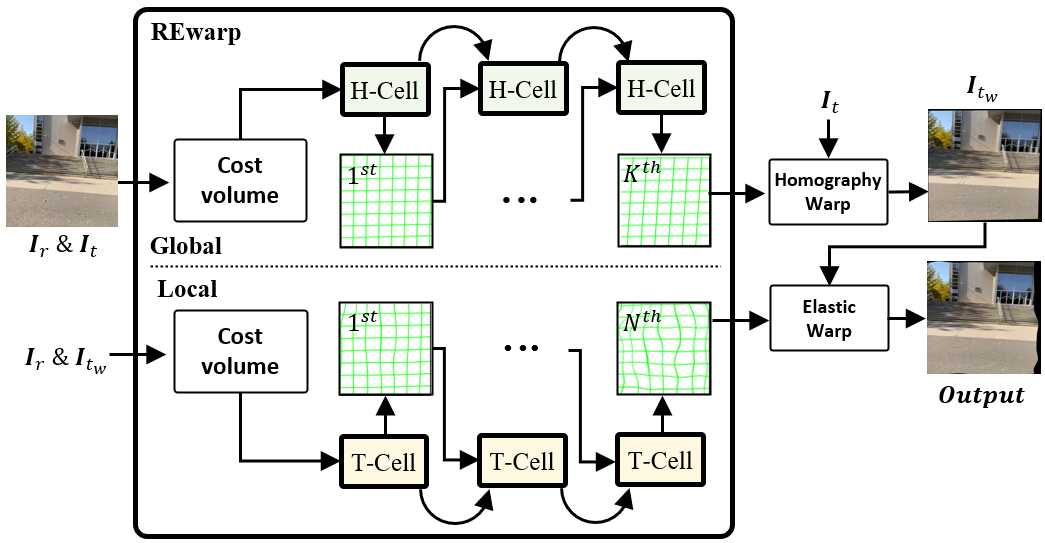}
    \vspace{-15pt}
    \caption{\textbf{Overview of our approach.} REwarp sequentially estimates homography and thin-plate spline with two recurrent neural networks. This combination enables elastic image alignment for parallax-tolerant image stitching.}
    \vspace{-10pt}
    \label{fig:teaser}
\end{figure}

On the other hand, the recent parallax-tolerant deep image stitchings \cite{kweon2021pixel, nie2023learning} are robust to few textured images owing to the CNNs with wide receptive fields and the prior of dense correspondences from correlation-based cost volumes \cite{xu2017accurate}. The methods are implemented by either a parametric combination (homography and \textit{Thin-plate Spline} (TPS) \cite{nie2023learning}) or a \textit{Warp Field} \cite{kweon2021pixel}. Because the TPS-based image deformation is a global warp, sometimes it shows limited correction of local misalignment under large parallax errors. In contrast, the flow-based image stitching shows warps with high flexibility for alignment of wide parallax. However, because the method is exposed to artifacts or holes (\cref{fig:TPS}), it requires modules to address the problems. Furthermore, the occasional discontinuity (\cref{fig:dirichlet}) of deep elastic warps between overlap and non-overlap regions requires seam masks \cite{kwatra2003graphcut, lin2016seagull, li2022automatic} to hide the unpleasant boundary.

\begin{figure*}[t]
    \footnotesize
    \begin{subfigure}[b]{0.49\textwidth}
        \begin{overpic}[height=1.0in, width=1.0in]       {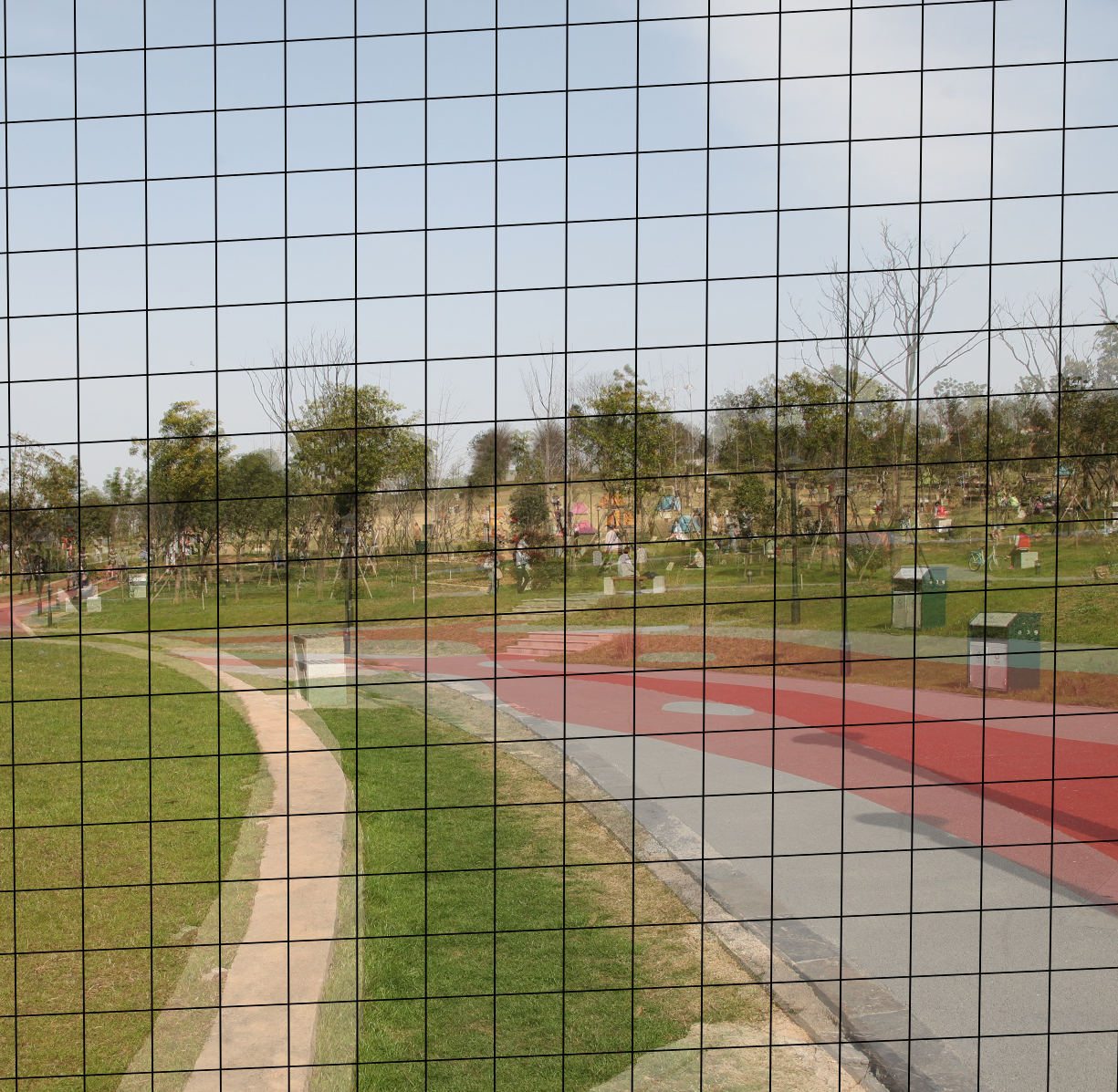}
            \put(5,85){G1}
        \end{overpic}
        \hspace{5pt}
        \begin{overpic}[height=1.0in, width=1.0in]       {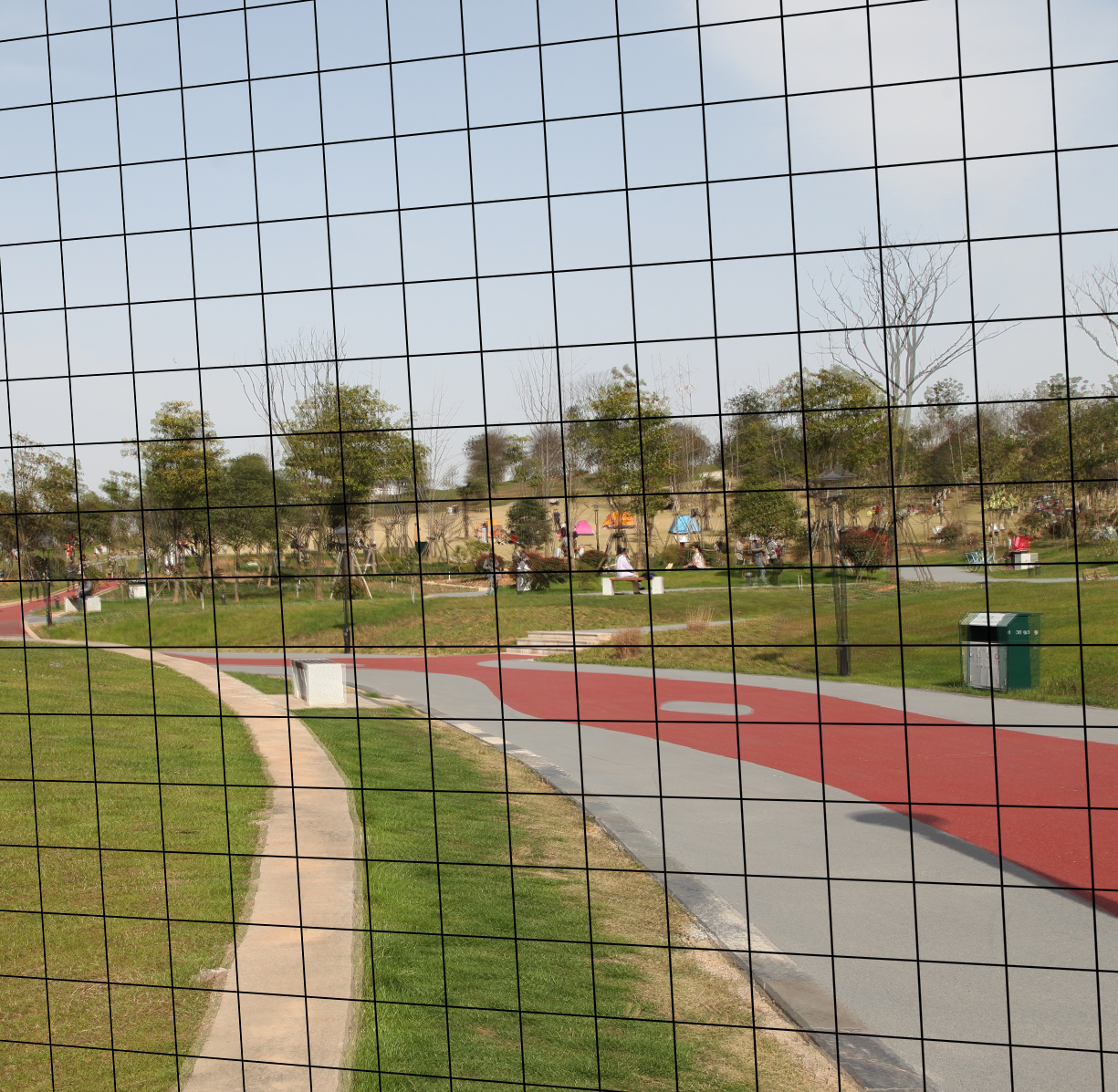}
            \put(5,85){G3}
        \end{overpic}
        \hspace{5pt}
        \begin{overpic}[height=1.0in, width=1.0in]       {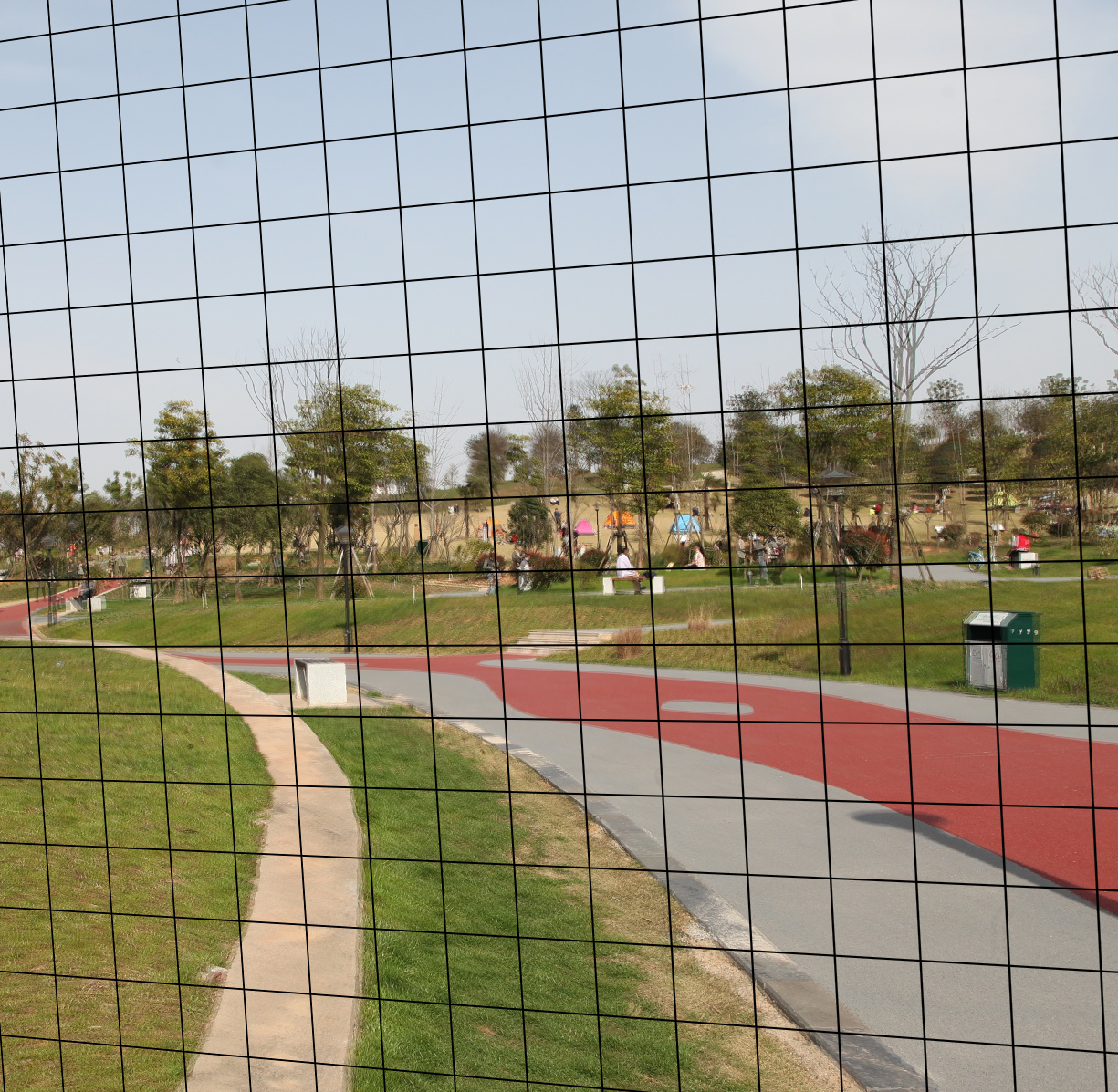}
            \put(5,85){G6}
        \end{overpic} \\
        \begin{overpic}[height=1.0in, width=1.0in]       {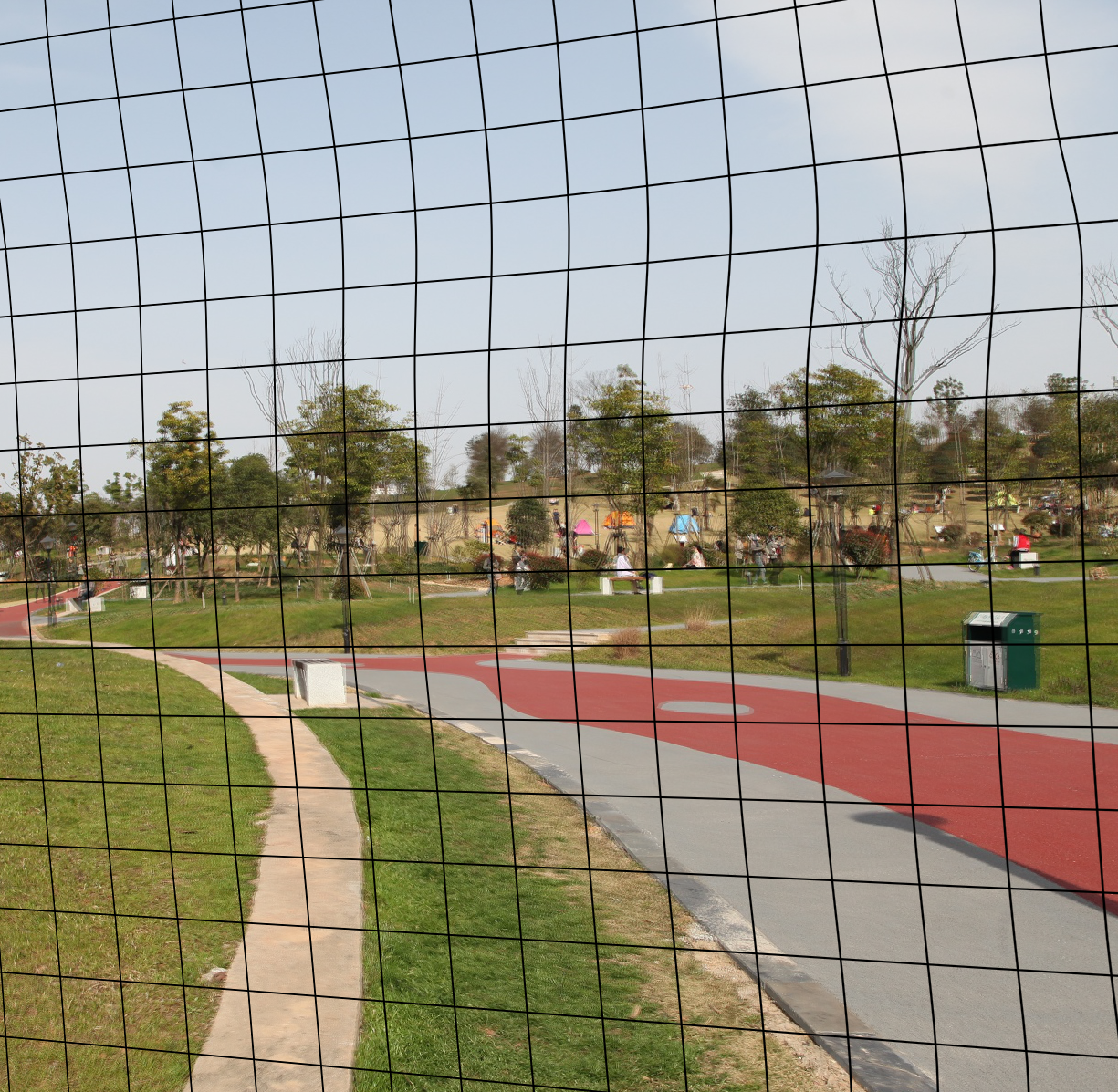}
            \put(5,85){L1}
        \end{overpic}
        \hspace{5pt}
        \begin{overpic}[height=1.0in, width=1.0in]       {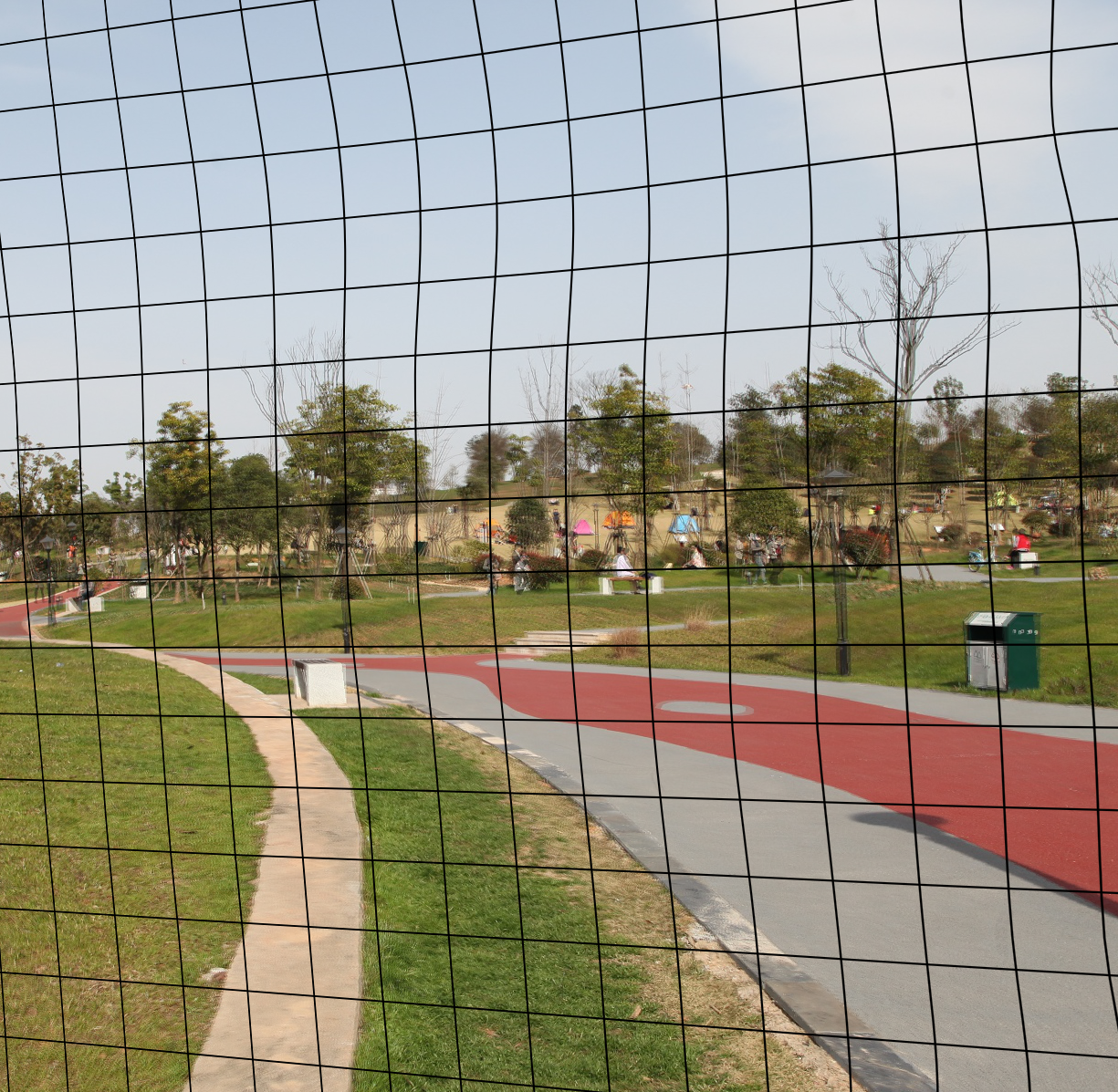}
            \put(5,85){L2}
        \end{overpic}
        \hspace{5pt}
        \begin{overpic}[height=1.0in, width=1.0in]       {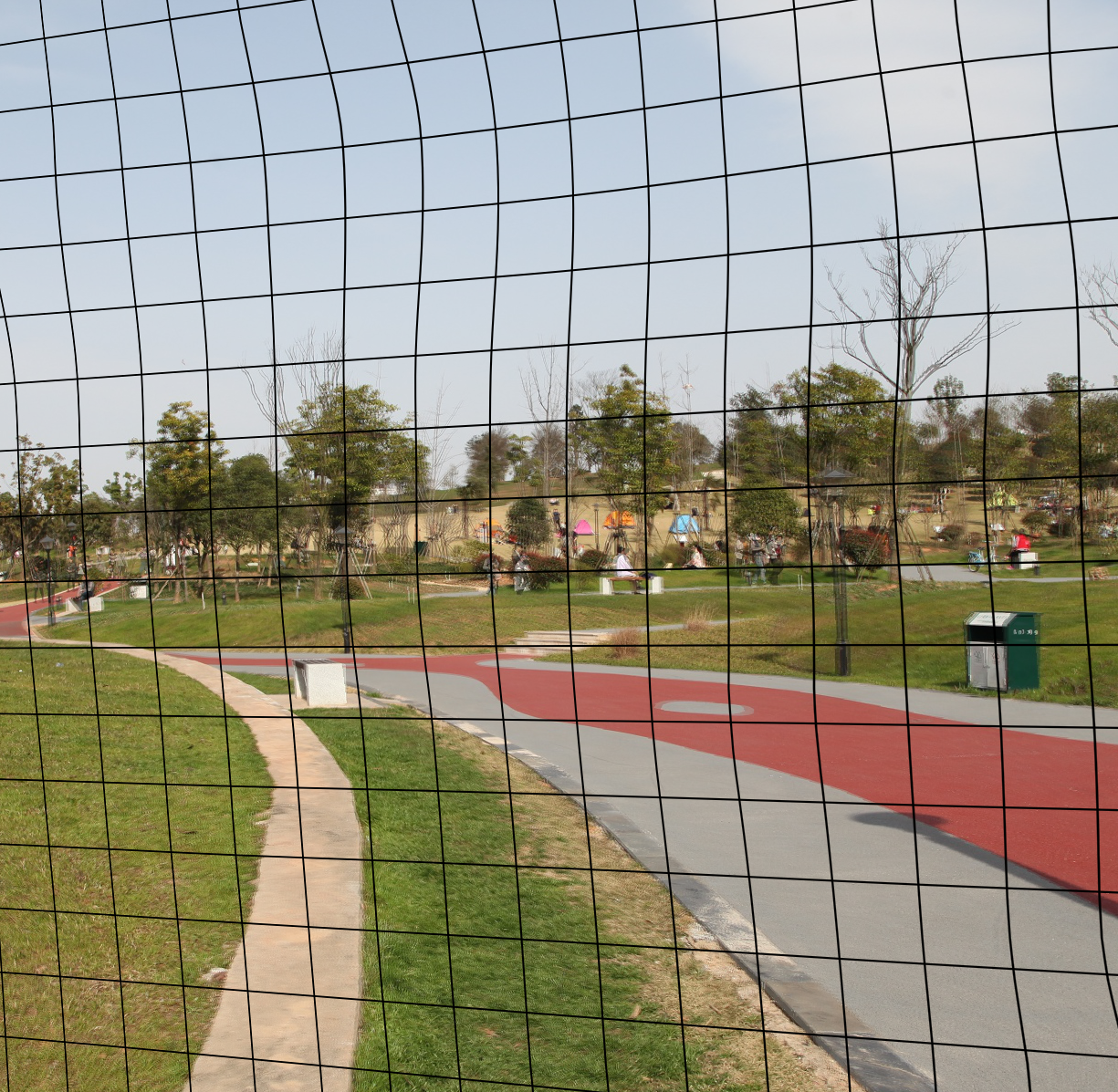}
            \put(5,85){L3}
        \end{overpic}
        \caption{\textbf{Observation on overlap region.} $1,2,4,6^{th}$ iterations of \\ global(top) and the residual(bottom) alignment}
    \end{subfigure}
    \begin{subfigure}[b]{0.49\textwidth}
        \centering
        \includegraphics[height=2.0in]{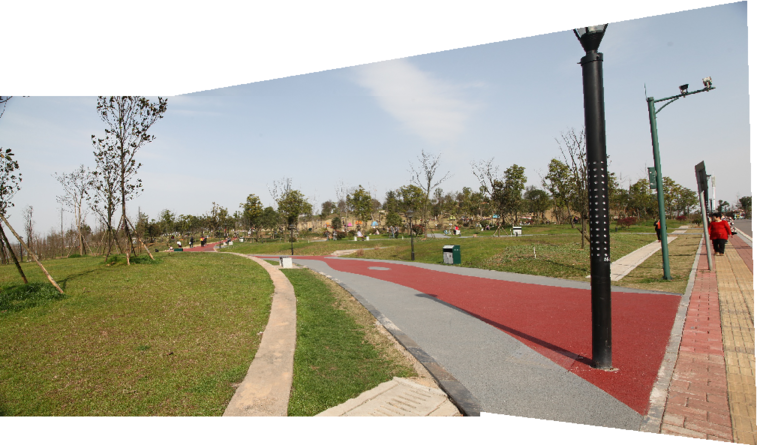}
        \caption{\textbf{Image Stitching with our approach.}}
    \end{subfigure}
    \vspace{-6pt}
    \caption{Visual Demonstration of our approach.}
    \vspace*{-10pt}
    \label{fig:visual_demo}
\end{figure*}

To address the problems, we suggest Recurrent Elastic warps (REwarp) with Dirichlet boundary condition that learns discontinuity-free residual elastic warps. For each inference of recurrent estimations, REwarp predicts an elastic warp parameterized by a homography and a Thin-plate spline. Our experiments show that REwarp addresses the generation of holes and discontinuity of learning-based elastic warps with remarkable image alignment.

\vspace{4pt}
In summary, our main contributions are as follows:
\begin{itemize}
\setlength{\leftmargin}{-.35in}
    \item We propose a novel deep image stitching that corrects residual misalignment and varying degrees of parallax errors using elastic warps.

    \item Our approach shows favorable performance without additional hole-filling and discontinuity correction modules required in the previous methods of deep elastic warp for image stitching.

    \item We investigate the computation costs of REwarp and our baselines to show that REwarp is a lightweight and failure-tolerant model. The properties will shed light on real-time deep image stitching. \\
\end{itemize}

\section{Related Work}
\noindent\textbf{Elastic Warp for image stitching}
As a global alignment by single homography lacks the consideration of local misalignment or parallax, there have been various studies and trials to overcome such a limitation. Gao \textit{et al.} \cite{gao2011constructing} suggested a method that determines two globally dominant frames and estimates dual homography. Lin \textit{et al.} \cite{lin2011smoothly} proposed smoothly varying affine fields. Zaragoza \textit{et al.} \cite{zaragoza2013projective} suggested moving Direct Linear Transformation (DLT) \cite{hartley2003multiple} to infer as-projective-as-possible (APAP) spatially weighted homography. Li \textit{et al.} \cite{li2017parallax} proposed robust elastic warps (robust ELA) that compute \textit{thin plate splinte} (TPS) to finely optimize pixel-wise deformations. Joo \textit{et al.} \cite{joo2015line} suggested a line-guided moving DLT for structure-preserving image stitching. Liao \textit{et al.} \cite{liao2019single} proposed single-perspective warps that leverage dual-feature (point + line) for structure-preserving image stitching. Lee \etal \cite{lee2020warping} introduced residual warping for large parallax alignment using multiple homography. Jia \textit{et al.} \cite{jia2021leveraging} suggested a line-point-consistence (LPC) constraint for advanced geometric structure-preserving image alignment. Kweon \etal \cite{kweon2021pixel} suggests Pixel-Wise warping Module (PWM) that learns warp field for wide-parallax image alignment. Recently, Nie \etal \cite{nie2023learning} proposed UDIS++ that contains a deep seamless compositor, a homography estimator, and a TPS estimator for deep image stitching. Despite the successful extends of deep image stitchings to elastic warps \cite{song2021end, kweon2021pixel, nie2023learning}, there are still challenging hurdles on completeness of aligned image composition. They struggle with discontinuity between overlap and non-overlap regions in target images. In this work, we address the problem and suggest novel deep residual warps for image stitching.

\begin{figure*}[t]
    \centering
    \includegraphics[width=0.90\linewidth]{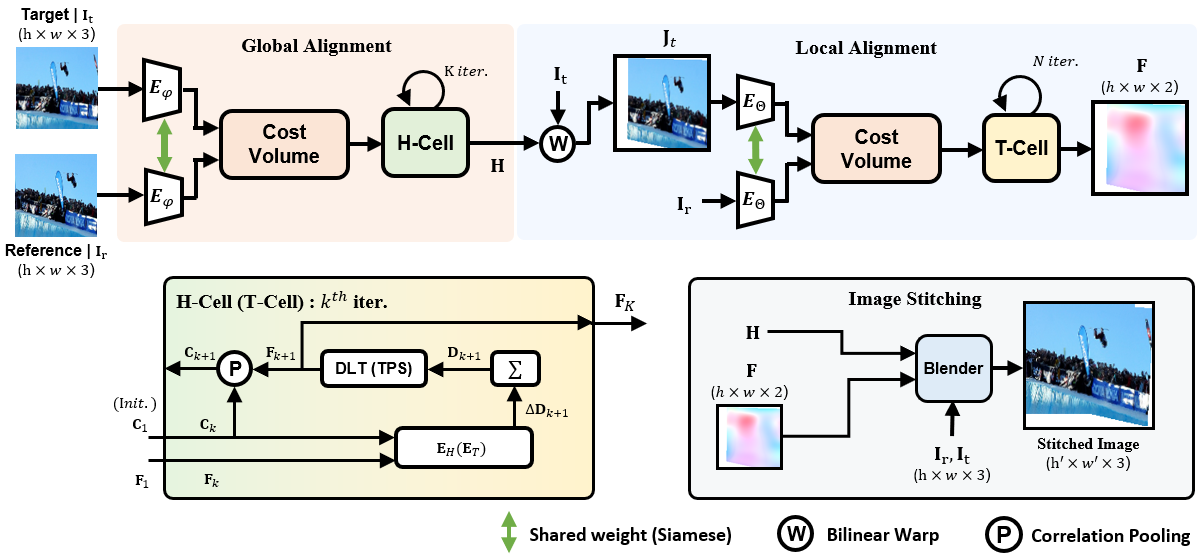}
    \vspace*{-7pt}
    \caption{\textbf{Architectural Overview of REwarp.} Our model consists of transformation estimators for global alignment and local residual alignment. After the estimation of both parametric warps, we ensemble them and composite an elastic warp field. In our H/T-Cell, an accumulator ($\Sigma$) sums estimated displacements ($\Delta \mathbf{D}_k$) from a regressor that consists of multi-layer CNNs ($\mathbf{E}_H$ or $\mathbf{E}_T$).}
    \label{fig:flowchart}
    \vspace*{-7pt}
\end{figure*}

\begin{figure}[t]
    \centering
    \includegraphics[width=2.7in]{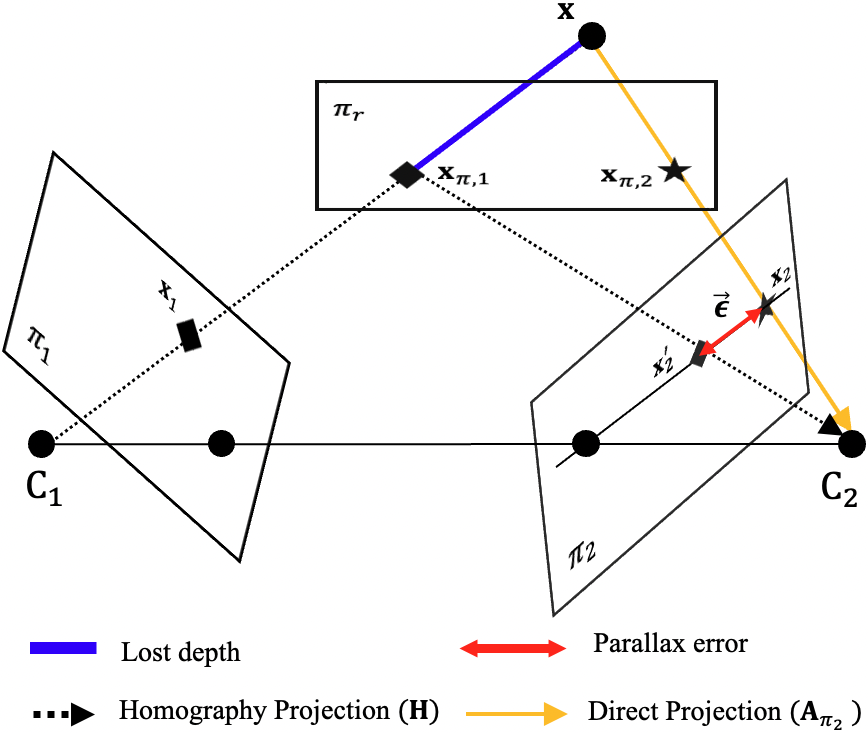}
    \vspace{-6pt}
    \caption{\textbf{A parallax error on point $\mathbf{x}_2$}. $\mathbf{\pi}$ denotes a projection plane, $\mathbf{C}$ is a view point, ${\mathbf{x}'}$ is a point projected by a homography.}
    \vspace*{-10pt}
    \label{fig:parallax}
\end{figure}

\section{Problem Formulation}
Parallax is a local misalignment caused by lost information on depth planes.  As depicted in \cref{fig:parallax}, the misalignment in projection plane $\boldsymbol{\pi}_2$ arises from the disparity between the homography projection ($\mathbf{H}:\mathbf{x}_1\mapsto{\mathbf{x}}_2'$) and the direct projection from real world to an image plane $\boldsymbol{\pi}_2$ ($\mathbf{A}_{\boldsymbol{\pi}_2}:\mathbf{x}\mapsto\mathbf{x}_2$)\cite{hartley2003multiple}. The mismatched depth ($\mathbf{x} - \mathbf{x}_{\boldsymbol{\pi}, 1}$) causes an intrinsic limitation to align such parallax error ($\vec{\boldsymbol{\epsilon}}$), which is represented as 
\vspace{-2pt}
\begin{align}
    \small
    \vec{\boldsymbol{\epsilon}} &= \mathbf{x}_2 -{\mathbf{x}}_2' =  \mathbf{A}_{\boldsymbol{\pi}_2}\mathbf{x}-\mathbf{H} \mathbf{x}_1, \\[-2pt] 
    &= \mathbf{A}_{\boldsymbol{\pi}_2}\mathbf{A}_{\boldsymbol{\pi}_1}^{-1}\mathbf{x}_1 -\mathbf{H} \mathbf{x}_1.
    \label{eq:woo_0}
\end{align}

To minimize the error $\vec{\boldsymbol{\epsilon}}$ on the plane, our network aims to predict a warp field $\widehat{\mathbf{{F}}}$ \cite{kweon2021pixel} to compensate $\vec{\boldsymbol{\epsilon}}$ and a homography $\widehat{\mathbf{H}}$ to globally align two images. Note that we denote $\widehat{(\cdot)}$ as an estimated warp used for image stitching. The overall formulations for alignment of parallax errors are the same as follows:
\begin{align}
    \small
    \min{\vec{\boldsymbol{\epsilon}}} \;\; & \equiv \;\; \min||\mathbf{x}_2 - (\widehat{\mathbf{H}}\cdot \mathbf{x}_1+\widehat{\mathbf{F}}[\mathbf{x}'_2])||, \\[-2pt] 
    & = \;\; \min||\mathbf{x}_2 - (\mathbf{x}'_2+\widehat{\mathbf{F}}[\mathbf{x}'_2])||, \nonumber \\
    & = \;\; \min||\mathbf{x}_2 - \widehat{\mathbf{x}}_2||, \nonumber \\[-20pt] \nonumber 
\end{align}
\noindent where $\mathbf{x}_2$ denotes an ideal point, $\mathbf{x}'_2$ is a point exposed to a prallax error $\vec{\boldsymbol{\epsilon}}$ by the homography projection. 

The estimated grid $\widehat{\mathbf{X}}_2:=\{\mathbf{x}_2|\mathbf{x}_2=\mathbf{x}'_2 + \vec{\boldsymbol{\epsilon}}\}$ is used for the image warping, which is defined as
\begin{align}
    \small
    W(\mathcal{I}, \mathcal{X}):=\mathcal{I}[\mathcal{X}],
\end{align}
where $\mathbf{I} \in \mathcal{I} \text{ and } \mathbf{X} \in \mathcal{X}$ is a set of images ($\in \mathbb{R}^{h\times w}$) and a warped grid of a stitched frame ($\in \mathbb{R}^{h'\times w'}$) respectively. To satisfy our formulations, REwarp estimates an accumulated displacement vector $\mathbf{D}^G$ and $\mathbf{D}^L$ for homography \cite{hartley2003multiple} and TPS computation, respectively. Specifically, we compute a warped grid $\widehat{\mathbf{X}}'_2$ using $\mathbf{D}^G$ and a warp field $\widehat{\mathbf{F}}\in\mathbb{R}^{h\times w\times 2}$ using $\mathbf{D}^L$ as 
\begin{gather}
    \small
    \widehat{\mathbf{X}}'_2 = \mathcal{S}_{\mathbf{V}}(\mathbf{D}^G)\cdot \mathbf{X}_1, \label{eq:dlt} \\
    \widehat{\mathbf{X}}_2 = \widehat{\mathbf{X}}'_2 + \mathcal{S}_{\mathbf{P}_r}(\mathbf{D}^L; \mathbf{X}^o),
\end{gather}

where $\mathcal{S}_\mathbf{V}$ and $\mathcal{S}_\mathbf{\mathbf{P}_r}$ denote a Direct Linear Transformation \cite{hartley2003multiple} and a warp field evaluation (\cref{subsec:wf}), respectively. $\mathbf{V} := \{0,h\}\times \{0,w\}$ is a vector containing four corner coordinates, $\mathbf{P}_r := \{0, \frac{1}{12}, ..., \frac{11}{12}\}^2 \in \mathbb{R}^{12\times 12}$ is a control point grid for $\mathbf{I}_r$ (or a pre-computed uniform grid). $\mathbf{X}^o$ means a coordinate set consisting of overlapping regions given by an estimated homography $\widehat{\mathbf{H}}=\mathcal{S}_{\mathbf{V}}(\mathbf{D}^G)$.


\begin{algorithm}[t]
\footnotesize
\caption{Operations in H-Cell and T-Cell.}\label{algorithm:cell}
\hspace*{2pt} \textbf{Input : }  $\mathbf{C}, k_\text{iter}$ \Comment{Cost Volume, Init., \# Iter.} \\
\hspace*{2pt} \textbf{Output : }  $\widehat{\mathbf{X}}$  \Comment{Elastic Grid}\\
\begin{algorithmic}[1]
    \Procedure{Cell}{$\mathbf{C}, k_{\text{iter}}$}
       \State  $\mathbf{F}_1,\; \mathbf{\delta}_1 \gets \vec{\mathbf{0}}$  
       \State $k \gets 1$ \\

       \While{$k\leq k_{\text{iter}}$}
          \State \#\# Prior Preparation
          \State $\mathbf{C}_k \gets f_c(\mathbf{C}; \mathbf{X}_k)$ \Comment{$f_c:=$ Pooling \cite{teed2020raft}} \label{costupdate}
          \State $\mathbf{s}_{k+1} = [\mathbf{C}_k, \mathbf{F}_k]$ \\
 
          \State \#\# Update \& Predictions
          \If {H-Cell}
              \State $\Delta\mathbf{\mathbf{D}}_{k+1}^G \gets \mathbf{E}_{H}(\mathbf{s}_{k+1})$ 
              \State $\mathbf{D}_{k+1}^G\gets \sum_{j=1}^{k+1} \Delta\mathbf{\mathbf{D}}_j^G $ 
              \label{accumulate}
              \State $\mathbf{X}_{k+1} \gets \mathcal{S}_{\mathbf{V}}(\mathbf{D}_{k+1}^G)$ \Comment{$\mathcal{S}_\mathbf{V}:=$\;DLT}\label{displacementupdate}
          \EndIf \\
          \If {T-Cell}
              \State $\Delta\mathbf{\mathbf{D}}_{k+1}^L \gets \mathbf{E}_{T}(\mathbf{s}_{k+1})$ 
              \State $\mathbf{D}_{k+1}^L\gets \sum_{j=1}^{k+1} \Delta\mathbf{\mathbf{D}}_j^L $
              \State $\mathbf{F}_{k+1}=\mathcal{S}_{\mathbf{P}_r}(\mathbf{D}^L)_{k+1}$ \Comment{$\mathcal{S}_{\mathbf{P}_r}:=\mathbf{F}$ Eval. (\cref{subsec:wf})}
              \State $\mathbf{X}_{k+1}[\mathbf{x}^o] = \mathbf{X}_k[\mathbf{x}^o] + \mathbf{F}_{k+1}[\mathbf{x}^o]$
          \EndIf
          \State $k\gets k+1$
       \EndWhile\label{cellwhile}
    \EndProcedure
\end{algorithmic}
\end{algorithm}

\section{Method}
REwarp consists of CNN encoders ($\mathbf{E}_\varphi, \mathbf{E}_\Theta$), an H-Cell ($f^G$) and a T-Cell ($f^L$). Both cells take in a 4D cost volume to estimate a homography and a TPS representation, respectively. Specifically, H-Cell and T-Cell predict a residual displacement vector $\Delta\mathbf{D}^G$ and $\Delta\mathbf{D}^L$ of four corners $\mathbf{V}$ \cite{detone2016deep} and a control point grid $\mathbf{P}_r$ \cite{bookstein1989principal}, respectively. The estimated residual vectors are accumulated and converged to the optimal vectors given $K$ (H-Cell) or $N$ (T-Cell) number of iterations as
\begin{gather} \label{eq:acc}
    \small
    \mathbf{D}^G = \sum_{k=1}^K \Delta \mathbf{D}^G_k, \;\; \mathbf{D}^L = \sum_{n=1}^N \Delta \mathbf{D}^L_n, \\
    \textrm{where} \;\; [\Delta\mathbf{D}^G_1, \dots, \Delta\mathbf{D}^G_K] = f^G(\mathbf{C}, K; \mathbf{I}_r, \mathbf{I}_t, \varphi), \nonumber \\
    [\Delta\mathbf{D}^L_1, \dots, \Delta\mathbf{D}^L_N] = \lambda \cdot f^L(\mathbf{C}, N; \mathbf{I}_r, \mathbf{J}_t, \Theta), \nonumber
\end{gather}
$\mathbf{J}_t$ denotes a target image warped by a homography $\mathbf{H}$ computed from $\mathbf{D}^G$. As shown in the above equation, a vector $\mathbf{D}^L$ for a control point grid is estimated from a globally aligned image $\mathbf{J}_t$ and a reference image $\mathbf{I}_r$ to correct the remaining parallax errors. More details and pipelines of our REwarp are provided in \cref{algorithm:cell}.


\subsection{Warp Field Evaluation} \label{subsec:wf}
We obtain a warp field ($\mathbf{F}$) via \textit{Thin-plate Spline} Transformation as
\begin{align}
    \small
    \mathbf{F} = \mathcal{S}_{\mathbf{P}_r}(\mathbf{D}^L) = \mathbf{X}_t - \mathbf{U},
\end{align}\vspace{4pt}
where $\mathbf{U}:= [0, h)\times [0, w)$ is an uniform grid, $\mathbf{X}_t \delequal \{\mathbf{x}_t|\mathbf{x}_t \in \mathbb{R}^2\}$ is a warped grid by TPS transformation.

\vspace{4pt}\noindent \textbf{TPS Transformation} An explicit representation TPS provides a warp field ($\mathbf{F}$) given control points $\mathbf{P}_r$ and $\mathbf{P}_t \in \mathbb{R}^{12\times 12}$. In this work, we assume that control points $\mathbf{P}_t$ in $\mathbf{I}_t$ are presented as $\widehat{\mathbf{P}}_t = \mathbf{P}_r + \mathbf{D}^L$ under the assumption that $\mathbf{P}_r$ is equal to a fixed uniform grid to ease the formulation for image alignment. Specifically, TPS calculates the affine weights ($\vec{v} :=[a, b, c]$) and kernel weights ($\vec{w}:=[w_1,w_2,\cdots,w_M]$) for each $x$ and $y$ axis as 
\begin{gather}
    \small
    \begin{bmatrix}
    \vec{w}_x & \vec{v}_x \\
    \vec{w}_y & \vec{v}_y
    \end{bmatrix} ^{T}
    = \mathbf{L}^{-1}
    \begin{bmatrix}
    \widehat{\mathbf{P}}_{tx}^T & 0 & 0 & 0 \\
    \widehat{\mathbf{P}}_{ty}^T & 0 & 0 & 0
    \end{bmatrix} ^{T},
    \label{eq:TPS}
\end{gather}
where $\mathbf{L}$ is a deterministic matrix corresponding the pairs of control points \cite{bookstein1989principal}, $[\widehat{\mathbf{P}}_{tx} ; \widehat{\mathbf{P}}_{ty}] = \widehat{\mathbf{P}}_t$. The TPS derives each coordinate of the target grid ($\mathbf{X}_t\delequal\{\mathbf{x}_t|\mathbf{x}_t\in \mathbb{R}^2\}$) with the calculated weights and reference grid $\mathbf{X}_r \delequal \{\mathbf{x}_r|\mathbf{x}_r\in\mathbb{R}^2\} \mbox{ in a range of } [0, h) \times [0, w)$, as
\begin{gather}
    \small
    \mathbf{x}_t =  \begin{bmatrix} {x}_{t}\\ {y}_{t}\end{bmatrix}=\begin{bmatrix}
        \vec{v}_x \cdot [1,\mathbf{x}_r] + \sum_m w_{x,m}B(||(\mathbf{x}_r - \mathbf{P}_{r,m}) ||)\\
        \vec{v}_y \cdot [1,\mathbf{x}_r] + \sum_m w_{y,m}B(||(\mathbf{x}_r - \mathbf{P}_{r,m}) ||)
    \end{bmatrix},\nonumber
\end{gather}
\noindent where $B(z)=z^2 logz^2$ is a radial basis function, $w_{A, m}$ denotes a $A \; (x \; or \; y)$ coordinate and $m^{th}$ kernel weight of M kernels.

\begin{figure}[t]
    \centering
    \stackunder[2pt]{
    \includegraphics[width=\linewidth]{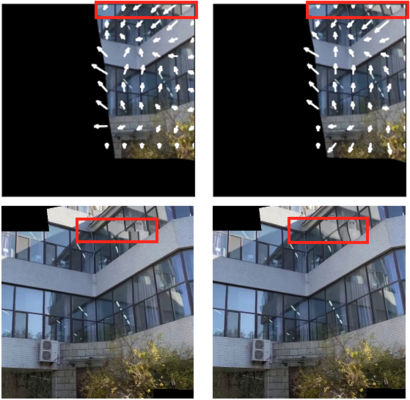}
    }{\textbf{w/o constraint} \qquad \qquad \quad \textbf{w/ constraint}}
    \caption{\textbf{Discontinuity Removal.} The white arrows denote displacement vectors of control points to be applied to overlap region warping. As in the red box, our constraint removes discontinuity by relaxing the displacements in the edge.}
    \vspace*{-10pt}
    \label{fig:Dirichlet}
\end{figure}

\begin{figure*}[t]
\footnotesize
\centering
    \begin{tikzpicture}[x=6cm, y=6cm, spy using outlines={every spy on node/.append style={smallwindow_w}}]
        \node[anchor=center] (FigA) at (0, 0) {\includegraphics[height=1.45in, width=2.0in]{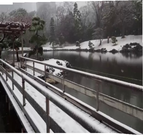}};
        \draw[red, line width=0.4mm] (-0.09,-0.11) rectangle (0.08,-0.29);
    \end{tikzpicture}
    \begin{tikzpicture}[x=6cm, y=6cm, spy using outlines={every spy on node/.append style={smallwindow_w}}]
        \node[anchor=center] (FigA) at (0,0) {\includegraphics[height=1.45in, width=1.45in]{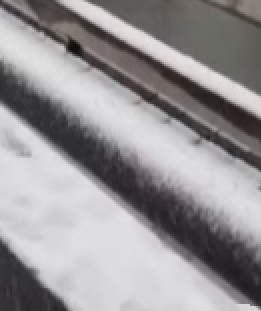}};
    \end{tikzpicture}
    \begin{tikzpicture}[x=6cm, y=6cm, spy using outlines={every spy on node/.append style={smallwindow_w}}]
        \node[anchor=center] (FigA) at (0,0) {\includegraphics[height=1.45in, width=1.45in]{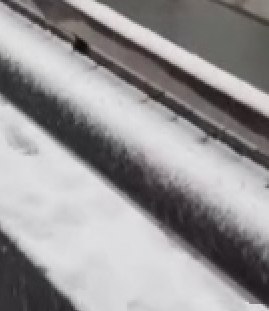}};
    \end{tikzpicture}
    \begin{tikzpicture}[x=6cm, y=6cm, spy using outlines={every spy on node/.append style={smallwindow_w}}]
        \node[anchor=center] (FigA) at (0,0) {\includegraphics[height=1.45in, width=1.45in]{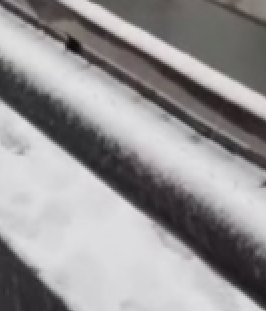}};
    \end{tikzpicture} \\
    \begin{tikzpicture}[x=6cm, y=6cm, spy using outlines={every spy on node/.append style={smallwindow_w}}]
        \node[anchor=center] (FigA) at (0, 0) {\includegraphics[height=1.45in, width=2.0in]{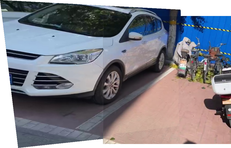}};
        \draw[red, line width=0.4mm] (-0.04,0.13) rectangle (0.11,0.28);
    \end{tikzpicture}
    \begin{tikzpicture}[x=6cm, y=6cm, spy using outlines={every spy on node/.append style={smallwindow_w}}]
        \node[anchor=center] (FigA) at (0,0) {\includegraphics[height=1.45in, width=1.45in]{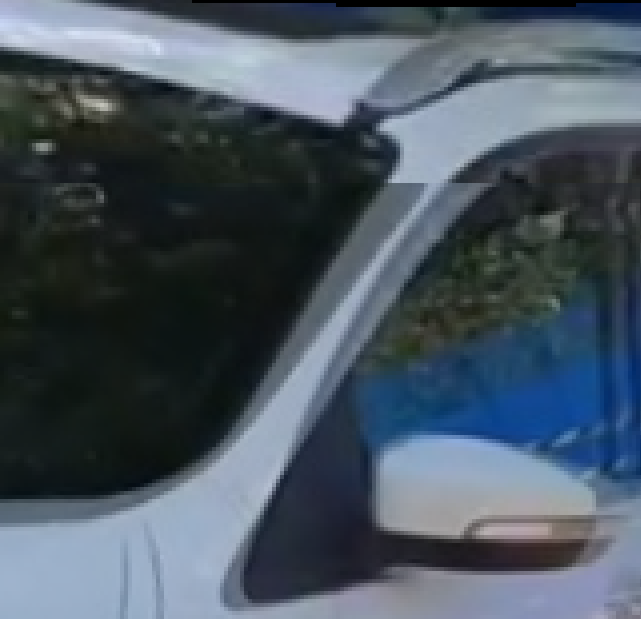}};
    \end{tikzpicture}
    \begin{tikzpicture}[x=6cm, y=6cm, spy using outlines={every spy on node/.append style={smallwindow_w}}]
        \node[anchor=center] (FigA) at (0,0) {\includegraphics[height=1.45in, width=1.45in]{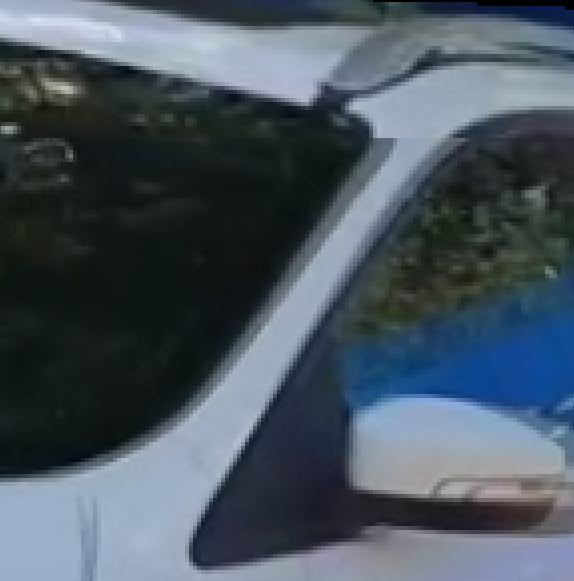}};
    \end{tikzpicture}
    \begin{tikzpicture}[x=6cm, y=6cm, spy using outlines={every spy on node/.append style={smallwindow_w}}]
        \node[anchor=center] (FigA) at (0,0) {\includegraphics[height=1.45in, width=1.45in]{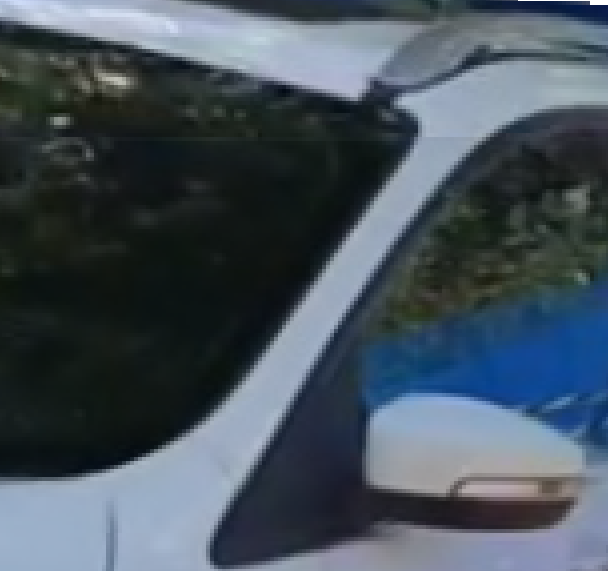}};
    \end{tikzpicture} \\
    \begin{tikzpicture}[x=6cm, y=6cm, spy using outlines={every spy on node/.append style={smallwindow_w}}]
        \node[anchor=center] (FigA) at (0, 0) {\includegraphics[height=1.45in, width=2.0in]{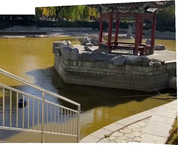}};
        \draw[red, line width=0.4mm] (-0.28,-0.22) rectangle (-0.13,-0.07);
    \end{tikzpicture}
    \begin{tikzpicture}[x=6cm, y=6cm, spy using outlines={every spy on node/.append style={smallwindow_w}}]
        \node[anchor=center] (FigA) at (0,0) {\includegraphics[height=1.45in, width=1.45in]{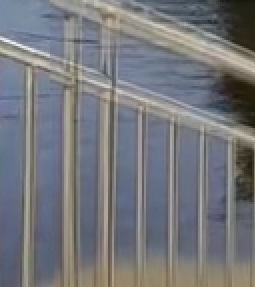}};
    \end{tikzpicture}
    \begin{tikzpicture}[x=6cm, y=6cm, spy using outlines={every spy on node/.append style={smallwindow_w}}]
        \node[anchor=center] (FigA) at (0,0) {\includegraphics[height=1.45in, width=1.45in]{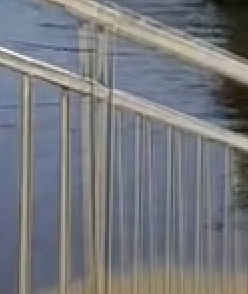}};
    \end{tikzpicture}
    \begin{tikzpicture}[x=6cm, y=6cm, spy using outlines={every spy on node/.append style={smallwindow_w}}]
        \node[anchor=center] (FigA) at (0,0) {\includegraphics[height=1.45in, width=1.45in]{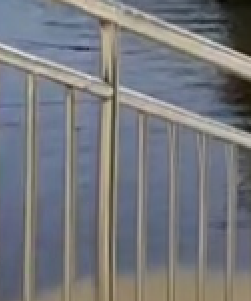}};
    \end{tikzpicture} \\
    \stackunder[2pt]{
        \begin{tikzpicture}[x=6cm, y=6cm, spy using outlines={every spy on node/.append style={smallwindow_w}}]
        \node[anchor=center] (FigA) at (0,0) {\includegraphics[height=1.45in, width=2.0in]{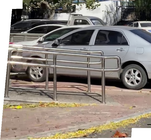}};
        \draw[red, line width=0.4mm] (0.12,-0.08) rectangle (0.27,0.07);
        \end{tikzpicture}
    }{Stitched Image}
    \stackunder[2pt]{
        \begin{tikzpicture}[x=6cm, y=6cm, spy using outlines={every spy on node/.append style={smallwindow_w}}]
        \node[anchor=center] (FigA) at (0,0) {\includegraphics[height=1.45in, width=1.45in]{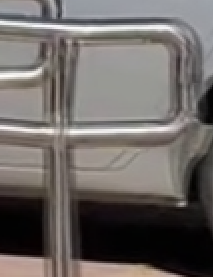}};
        \end{tikzpicture}
    }{Robust ELA \cite{li2017parallax}}
    \hspace{-6pt}
    \stackunder[2pt]{
        \begin{tikzpicture}[x=6cm, y=6cm, spy using outlines={every spy on node/.append style={smallwindow_w}}]
        \node[anchor=center] (FigA) at (0,0) {\includegraphics[height=1.45in, width=1.45in]{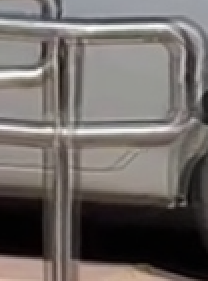}};
        \end{tikzpicture}
    }{LPC \cite{jia2021leveraging}}
    \hspace{-7pt}
    \stackunder[2pt]{
        \begin{tikzpicture}[x=6cm, y=6cm, spy using outlines={every spy on node/.append style={smallwindow_w}}]
        \node[anchor=center] (FigA) at (0,0) {\includegraphics[height=1.45in, width=1.45in]{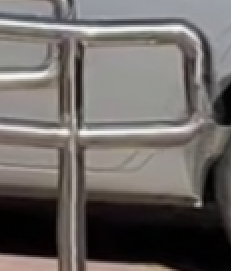}};
        \end{tikzpicture}
    }{REwarp (\textit{ours})}
    \vspace*{-5pt}
    \caption{\textbf{Qualitative comparisons to the other methods.}}
    \vspace*{-10pt}
\label{fig:Qual}
\end{figure*}
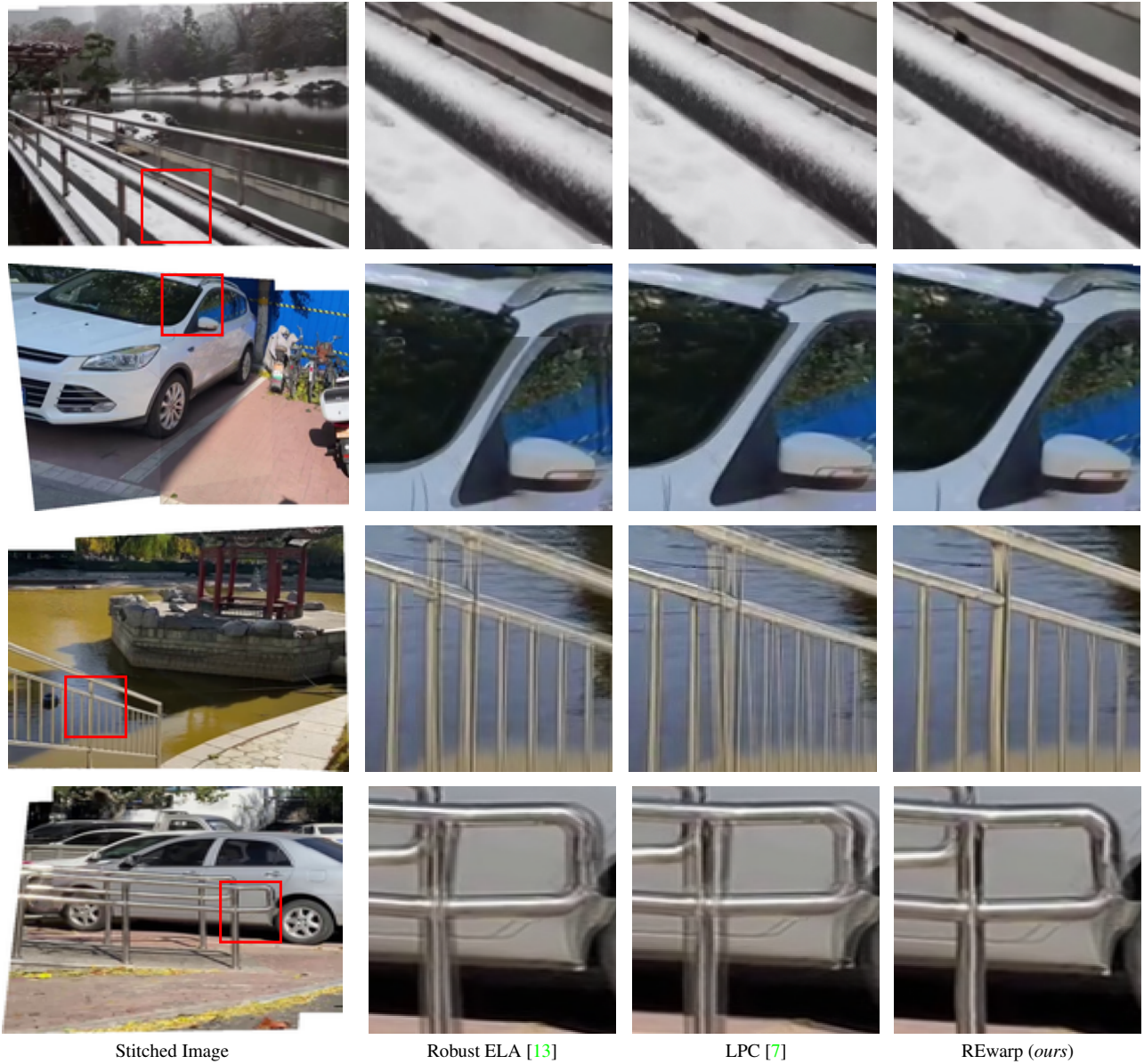

\noindent\textbf{Dirichlet Boundary Condition}
Though the recent remarkable deep parallax-tolerant warps \cite{kweon2021pixel, nie2023learning} can align paired images with wide-parallax errors, the methods struggle with discontinuity between overlap and non-overlap regions. To address the problem, we suggest a \textit{Dirichlet Boundary Condition} for TPS-based deep image stitching. Specifically, we design REwarp to fix the edge of the control point grid ($\mathbf{P}_t \in \mathbb{R}^{12\times 12}$) and leverage the estimated displacement vectors ($\Delta \mathbf{D}^L \in \mathbb{R}^{10\times 10}$). After that, we plug the vectors into the prepared control point grid. As shown in \cref{fig:Dirichlet}, this constraint prevents REwarp from inconsistent discontinuity because TPS coefficients in \cref{eq:TPS} provide almost no warp in the edges of the overlap region owing to the property of radial basis function. 

\subsection{Training}
 \noindent \textbf{H-Cell} We minimize photometric errors of overlapping region as
\vspace{-5pt}
\begin{gather}
    \small
    (\mathcal{L}_H)~:~\min_{\Theta_G} \sum_{k=1}^K\alpha^{K-k} \cdot \sum_{j\in\mathcal{J}_k} \; \|\mathbf{I}_r[\mathbf{x}^j_r] - \mathbf{I}_t[\mathbf{x}^j_t]\|_1,
    \label{eq:global}
\end{gather} \vspace{-5pt} \\
\noindent where $\Theta_G$ denotes parameters in H-Cell. $\mathcal{J}_k$ denotes a valid region coordinate in the $k^{th}$ iteration's overlap region ($\mathcal{J} \delequal \{j|j\in\mathbb{R}^2\}$), $\mathbf{x}^j_A$ is an overlap region coordinate in the plane of A. $\alpha$ is a weight $(=0.85)$. $K$ is the total iteration number of the H-Cell set as 6 in our implementation.

\begin{table*}[t]
    \small
    \centering
    \setlength{\tabcolsep}{1.5pt}
    \renewcommand{\arraystretch}{1.2}
    \begin{tabular}{
        c|>{\centering\arraybackslash}p{0.24\columnwidth}
        >{\centering\arraybackslash}p{0.24\columnwidth}
        >{\centering\arraybackslash}p{0.24\columnwidth}
        |>{\centering\arraybackslash}p{0.22\columnwidth}
        >{\centering\arraybackslash}p{0.22\columnwidth}
        >{\centering\arraybackslash}p{0.22\columnwidth}
    }
        \Xhline{2\arrayrulewidth}
         & \multicolumn{3}{c|}{Overlap Ratio} & \multirow{2}{*}{Average ($\uparrow$)} & \multirow{2}{*}{Failures ($\downarrow$)} & Time \\
         & $\sim30\% \; (\uparrow)$ & $31\sim60\% \; (\uparrow)$ & $61\%\sim \; (\uparrow)$ & & & (ms) \\
        \hline
        SIFT \cite{lowe2004distinctive} + RANSAC \cite{fischler1981random} & 18.32 & 21.68 & 22.30 & 21.48 & 1.27\% & 111 \\
        UDIS \cite{nie2021unsupervised} & 19.61 & 20.15 & 19.88 & 19.97 & \textcolor{red}{0\%} & - \\
        IHN \cite{cao2022iterative} & 20.09 & 21.73 & 23.27 & 22.99 & \textcolor{red}{0\%} & \textcolor{red}{38} \\
        APAP\cite{zaragoza2013projective} & 21.28 & 22.30 & 23.54 & 22.69 & 12.30\% & 574 \\
        SPW\cite{liao2019single} & 20.74 & 21.71 & 22.45 & 21.95 & 85.08\% & 383 \\
        LPC\cite{jia2021leveraging} & 17.97 & 21.04 & 21.59 & 20.82 & 42.13\% & 1395 \\
        Robust ELA\cite{li2017parallax} & \textcolor{blue}{21.84} & \textcolor{blue}{22.91} & \textcolor{blue}{24.29} & \textcolor{blue}{23.48} & \textcolor{blue}{0.72\%} & 79 \\
        REwarp (\textit{ours}) & \textcolor{red}{22.11} & \textcolor{red}{24.55} & \textcolor{red}{26.08} & \textcolor{red}{24.84} & \textcolor{red}{0\%} & \textcolor{blue}{50} \\
        \Xhline{2\arrayrulewidth}
    \end{tabular}
    \vspace{-4pt}
    \caption{\textbf{Quantitative Comparisons on Image Alignment.} \textcolor{red}{Red} (or \textcolor{blue}{Blue}) indicates the best (or second-best) performance.}
    \label{tab:Quan}
\end{table*}

\begin{table}[t]
    \centering
    \small
    \begin{subtable}[t]{\linewidth}
        \centering
        \begin{tabular}{
            c
            >{\centering\arraybackslash}p{0.22\linewidth}
            >{\centering\arraybackslash}p{0.22\linewidth}
            >{\centering\arraybackslash}p{0.22\linewidth}
        }
            \Xhline{2\arrayrulewidth}
            \multirow{2}{*}{Method} & Iteration & mPSNR & Time \\
            & (Num.) & (dB) & (ms) \\
            \hline
            
            \multirow{4}{*}{H-Cell} & 1 & 19.15 & 32 \\            
             & 3 & 23.81 & 36 \\
             & 6 & \textbf{24.84} & 50 \\
             & 10 & 24.79 & 67 \\
            \Xhline{2\arrayrulewidth}
            \multirow{4}{*}{T-Cell} & 1 & 24.49 & 47 \\            
             & 3 & \textbf{24.84} & 50 \\
             & 6 & 24.80 & 61 \\
             & 10 & 22.75 & 82 \\           
            \Xhline{2\arrayrulewidth}
        \end{tabular}
        \subcaption{Number of Iterations.}
        \label{subtab:iteration}
        \vspace{10pt}
        \begin{tabular}{
            >{\centering\arraybackslash}p{0.28\linewidth}
            >{\centering\arraybackslash}p{0.28\linewidth}
            >{\centering\arraybackslash}p{0.28\linewidth}
        }
            \Xhline{2\arrayrulewidth}
            Method & mPSNR (dB) & Parameter (M) \\
            \hline
            $\mathbf{H}$ & 22.31 & 1.0 \\
            $\mathbf{H}+\mathbf{F}$ & 24.43 & 1.7 \\
            $\mathbf{H}+$TPS (Ours) & \textbf{24.84} & 1.8 \\
            \Xhline{2\arrayrulewidth}
            \end{tabular}
        \subcaption{Implementation approach for Elastic Warp.}
        \label{subtab:elwarp}
    \end{subtable}
    \vspace{-5pt}
    \caption{\textbf{Ablation Study on Method specifications.}}
    \vspace{-10pt}
    \label{tab:ablation}
\end{table}

\noindent \textbf{T-Cell} We freeze H-Cell and train T-Cell to make REwarp strictly focus on learning the correction of residual parallax. Similar to \cref{eq:global}, we minimize L1 Loss as
\vspace{-2pt}
\begin{gather} \label{eq:elastic}
    \small
    (\mathcal{L}_T)~:~\min_{\Theta_L} \sum_{n=1}^N\alpha^{N-n} \cdot \sum_{q \in\mathcal{Q}_n}\|
 (\mathbf{I}_r[\mathbf{x}^q_r]-\mathbf{J}_t[\mathbf{x}^q_t])\|_1, \nonumber \\
    \textrm{where} \quad \mathbf{J}_t = W(\mathbf{I}_t,\;\widehat{\mathbf{H}}\cdot \mathbf{X}_t),\nonumber
\end{gather}

\noindent where $\mathbf{X}_t\in\mathbb{R}^{h\times w\times 2}$ is a grid of a target image, $\mathcal{Q}_n \delequal \{q|q\in\mathbb{R}^2\}$ denotes a query set of $n^{th}$ overlapping region, $N$ is the total number of iterations for T-Cell that is set as 3 in our implementation. The optimization of TPS coefficient estimators is guided by the photometric loss, i.e., the criteria is to maximize the correlation of overlapped regions. 

\section{Experiment} \label{sec:exp}
In our experiments, we use the UDIS-D \cite{nie2021unsupervised} dataset that includes real images exposed to various degrees of parallax errors. We follow the splitting method of the pioneering work \cite{nie2021unsupervised}.

\subsection{Implementation Details}
To optimize REwarp, we apply 2-stage learning. First, we train H-Cell during 100 epochs and 8 mini-batch. After then, we set the epoch for T-Cell as 100 with 8 mini-batch. \\
\noindent\textbf{Common Configurations}
We use AdamW optimizer \cite{loshchilov2017decoupled} with $\beta_1=0.9$ and $\beta_2=0.999$. The learning rates are initialized as $1\times10^{-4}$ and exponentially decayed by a factor of 0.98 for every epoch. We employ a $12\times 12$ control point grid. Note that because we constrain the grid with Dirichlet Boundary Condition, REwarp estimates $11\times 11$ grid and adds it to the $12\times 12$ zero-initialized grid as described in \cref{subsec:wf}.

\begin{figure*}[t]
    \centering
    \begin{tikzpicture}[x=6cm, y=6cm, spy using outlines={every spy on node/.append style={smallwindow_w}}]
    \node[anchor=center] (FigA) at (0,0) {\includegraphics[height=0.30\linewidth, width=0.45\linewidth]{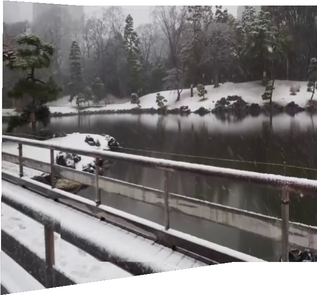}};
    \draw[red, line width=0.50mm] (-0.45,-0.38) rectangle (-0.36,-0.06);
    \draw[blue, line width=0.50mm] (-0.33,-0.38) rectangle (-0.03,-0.27);
    \end{tikzpicture}
    \hspace{10pt}
    \begin{tikzpicture}[x=6cm, y=6cm, spy using outlines={every spy on node/.append style={smallwindow_w}}]
    \node[anchor=center] (FigA) at (0,0) {\includegraphics[height=0.30\linewidth, width=0.45\linewidth]{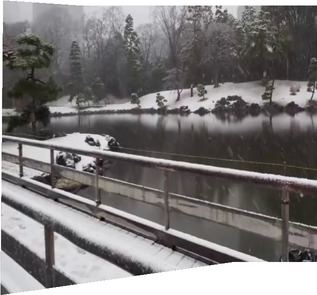}};
    \draw[red, line width=0.50mm] (-0.45,-0.38) rectangle (-0.36,-0.06);
    \draw[blue, line width=0.50mm] (-0.33,-0.38) rectangle (-0.03,-0.27);
    \end{tikzpicture}
    \stackunder[2pt]{
        \begin{tikzpicture}[x=6cm, y=6cm, spy using outlines={every spy on node/.append style={smallwindow_w}}]
        \node[anchor=center] (FigA) at (0,0) {\includegraphics[height=0.30\linewidth, width=0.45\linewidth]{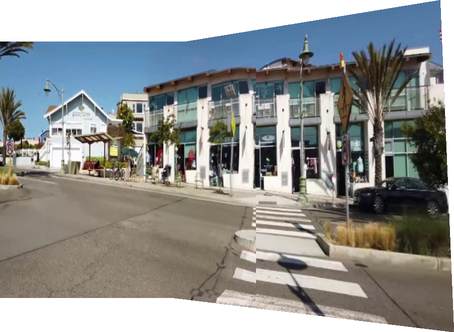}};
        \draw[red, line width=0.50mm] (0.03,-0.33) rectangle (0.13,0.28);
        \end{tikzpicture}
    }{\textbf{w/o Constraint}}
    \stackunder[2pt]{
        \begin{tikzpicture}[x=6cm, y=6cm, spy using outlines={every spy on node/.append style={smallwindow_w}}]
        \node[anchor=center] (FigA) at (0,0) {\includegraphics[height=0.30\linewidth, width=0.45\linewidth]{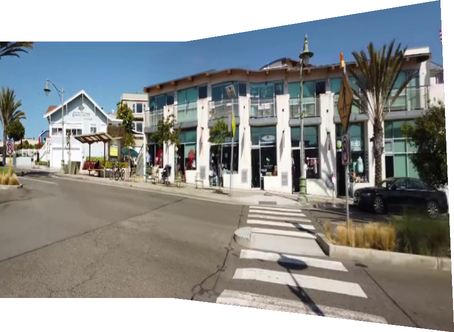}};
        \draw[red, line width=0.50mm] (0.03,-0.33) rectangle (0.13,0.28);
        \end{tikzpicture}
    }{\textbf{w/ Constraint}}
    \caption{\textbf{Ablation Study on Dirichlet Boundary Condition.}}
    \label{fig:dirichlet}
    \vspace{-10pt}
\end{figure*}

\subsection{Evaluation}
\noindent\textbf{Qualitative Comparison}
\cref{fig:Qual} shows the comparisons of REwarp to the previous parallax-tolerant stitchings \cite{li2017parallax, jia2021leveraging}. The first column contains REwarp's stitched results. The zoom-in comparisons among Robust ELA, LPC, and REwarp are provided in the second to the last columns, respectively. For fair comparisons, we average overlapping regions to composite stitched images. As shown in the figures, we notice that REwarp resolves the hole generations and discontinuity between overlap and non-overlap regions without any additional prevention methods \cite{kweon2021pixel}. Under the natural scenes like the first, and last rows, the other baselines with REwarp show reasonable alignment. However, we see that our approach empirically provides more stable results. In this experiment, while Robust ELA and LPC show frequent failures under the images with few textures like the third row (stair), REwarp shows relaxed favorable results. The experiences support the performance gains in the next quantitative comparisons.

\noindent\textbf{Quantitative Comparison}
In \cref{tab:Quan}, we report photometric errors using mPSNR (valid region-only PSNR) considering that measurement with PSNR contains the contribution of invalid black regions. The reported times are measured by averaging repeated 100 inferences under NVIDIA RTX 3090 24GB (or AMD Ryzen7 4800H). 
The `Failures' column provides the ratio of evaluation failures including no overlapping regions and huge image resolutions caused by unreasonably estimated warps. To observe the performances with the difficulty of image stitching, we additionally investigate the overlapping regions. Note that stitching under a low overlap ratio is a more challenging configuration compared to a high ratio. In the provided table, we see that the computation of REwarp is competitive for the existing image stitchings. In addition, the comparison between IHN and REwarp implies the effectiveness of our warps showing a 1.85dB performance gain. The comparison between Robust ELA and REwarp informs that the residual learning for misalign correction provides remarkable performance gains. On the other hand, the failure ratios of structure-preserving image stitching \cite{liao2019single, jia2021leveraging} indicate that their constraints may boost failure cases and cause harm to the overall aligning performances.

\subsection{Ablation Study}
We explore the contributions of three methods of REwarp including iterative estimation, implementation approach for elastic warps, and boundary constraint. As a default model configuration, we fix the number of H-Cell and T-Cell iterations as 6 and 3, respectively.

\noindent\textbf{Iterative Estimation}
In \cref{subtab:iteration}, we investigate the relationship between the number of iterations and model performance. As shown in the table, H-Cell stably converges to a global optimum under model capacity while T-Cell shows divergence at 10 iterations. The result implies that the warp field's high degree provides limited model capacity for the extreme number of iterations.

\noindent\textbf{Implementation Approach for Elastic warp}
In \cref{subtab:elwarp}, we explore methods that can be used for the implementation of elastic warp. As an approach for $\mathbf{H}$, we use our H-Cell. To prepare the estimator for $\mathbf{F}$, we simply revise the regressor of T-Cell to a regressor estimating $\mathbf{F}\in\mathbb{R}^{h\times w}$. Our investigation shows the effectiveness of the TPS-based image alignment achieving a 0.41dB mPSNR gain compared to the flow-based warp.

\noindent\textbf{Boundary Constraint} For empirical observation that our constraint using Dirichlet Boundary Condition prevents REwarp from the discontinuity between overlap and non-overlap regions, we explore  qualitative comparisons in \cref{fig:dirichlet}. In this study, we use linear blending considering that the guarantee of continuity makes the stitched image exposed to misalignment. As shown in the left column of \cref{fig:dirichlet}, the flow addition in the overlap region of the target image ($\mathbf{I}_t$) causes discontinuity. However, our simple constraint resolves such a problem.

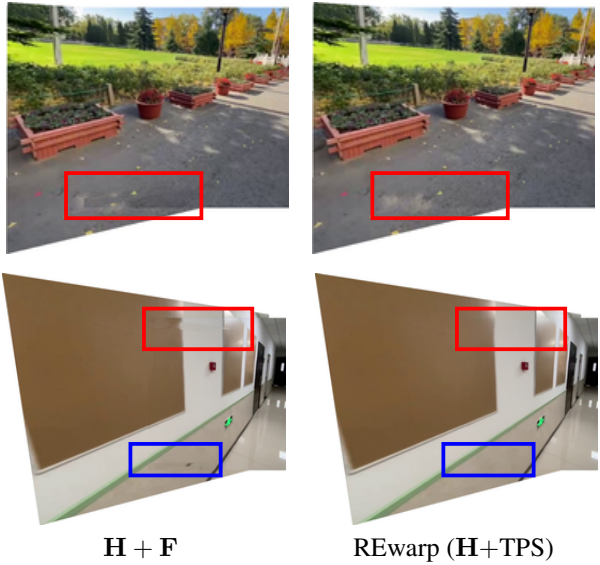
\begin{figure}[t]
    \centering
    \begin{tikzpicture}[x=6cm, y=6cm, spy using outlines={every spy on node/.append style={smallwindow_w}}]
    \node[anchor=center] (FigA) at (0,0) {\includegraphics[height=0.40\linewidth, width=0.45\linewidth]{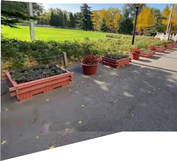}};
    \draw[red, line width=0.50mm] (-0.18,-0.10) rectangle (0.12,-0.20);
    \end{tikzpicture}
    \begin{tikzpicture}[x=6cm, y=6cm, spy using outlines={every spy on node/.append style={smallwindow_w}}]
    \node[anchor=center] (FigA) at (0,0) {\includegraphics[height=0.40\linewidth, width=0.45\linewidth]{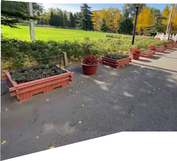}};
    \draw[red, line width=0.50mm] (-0.18,-0.10) rectangle (0.12,-0.20);
    \end{tikzpicture}
    \vspace{5pt}
    \stackunder[2pt]{
        \begin{tikzpicture}[x=6cm, y=6cm, spy using outlines={every spy on node/.append style={smallwindow_w}}]
        \node[anchor=center] (FigA) at (0,0) {\includegraphics[height=0.40\linewidth, width=0.45\linewidth]{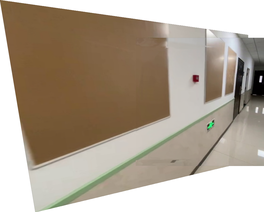}};
        \draw[red, line width=0.50mm] (0.00,0.11) rectangle (0.24,0.20);
        \draw[blue, line width=0.50mm] (-0.03,-0.10) rectangle (0.17,-0.17);
        \end{tikzpicture}}{$\mathbf{H}+\mathbf{F}$}
        \stackunder[2pt]{
        \begin{tikzpicture}[x=6cm, y=6cm, spy using outlines={every spy on node/.append style={smallwindow_w}}]
        \node[anchor=center] (FigA) at (0,0) {\includegraphics[height=0.40\linewidth, width=0.45\linewidth]{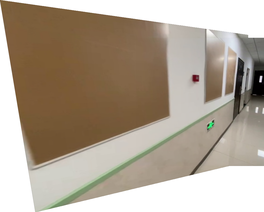}};
        \draw[red, line width=0.50mm] (0.00,0.11) rectangle (0.24,0.20);
        \draw[blue, line width=0.50mm] (-0.03,-0.10) rectangle (0.17,-0.17);
        \end{tikzpicture}}{REwarp ($\mathbf{H}+$TPS)}
    \vspace{-10pt}
    \caption{\textbf{Qualitative Comparison of Flow-based and TPS-based deep image stitching.}}
    \label{fig:TPS}
    \vspace{-13pt}
\end{figure}

\section{Discussion}
\noindent \textbf{Property of Flow and TPS-based Alignment} Although flow-based deep image stitching enables highly flexible warps, it struggles with artifacts in low-frequency textures and discontinuity between overlap and non-overlap regions as in \cref{fig:TPS}. In contrast, the TPS-based method provides relatively rigid flexibility compared to the method. Despite the property, TPS's non-linear and smooth warps with our boundary condition are free from the limitations of flow-based stitching satisfying most alignments for parallax. Though our constraint resolves the discontinuity, it may cause visually unnatural image stitching. Future works addressing the problem will be promising.

\begin{figure}[t]
    \hspace{-1.5mm}
    \centering
    \stackunder[2pt]{
        \begin{tikzpicture}[x=6cm, y=6cm, spy using outlines={every spy on node/.append style={smallwindow}}]
        \node[anchor=south] (FigA) at (0,0) {\includegraphics[width=0.95\linewidth]{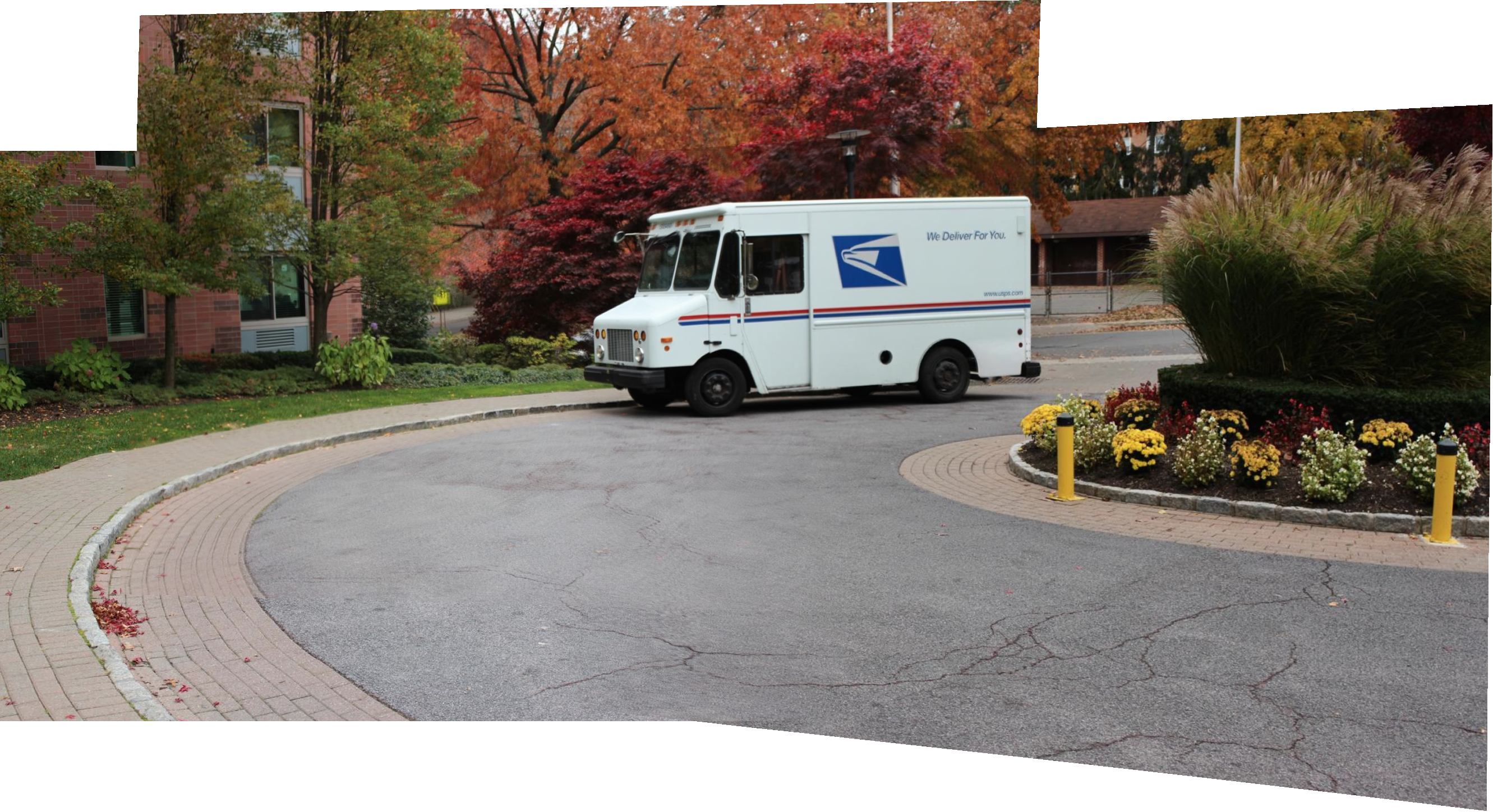}};
        \spy [closeup_w,magnification=2.5] on ($(FigA)+(-0.06,0.133)$) 
        in node[smallwindow,anchor=south] at ($(FigA.south) + (-0.06,0.115)$);
        \spy [closeup_r,magnification=2.5] on ($(FigA)+(-0.46,-0.15)$)
        in node[smallwindow,anchor=south] at ($(FigA.south) + (-0.46,0.35)$);
        \spy [closeup_r,magnification=2.0] on ($(FigA)+(0.110,0.170)$)
        in node[smallwindow,anchor=south] at ($(FigA.south) + (0.40,0.115)$);
        \end{tikzpicture}
    }{LPC \cite{jia2021leveraging}}
    \vspace{6pt}
    
    \stackunder[2pt]{
        \begin{tikzpicture}[x=6cm, y=6cm, spy using outlines={every spy on node/.append style={smallwindow}}]
        \node[anchor=south] (FigA) at (0,0) {\includegraphics[width=0.95\linewidth]{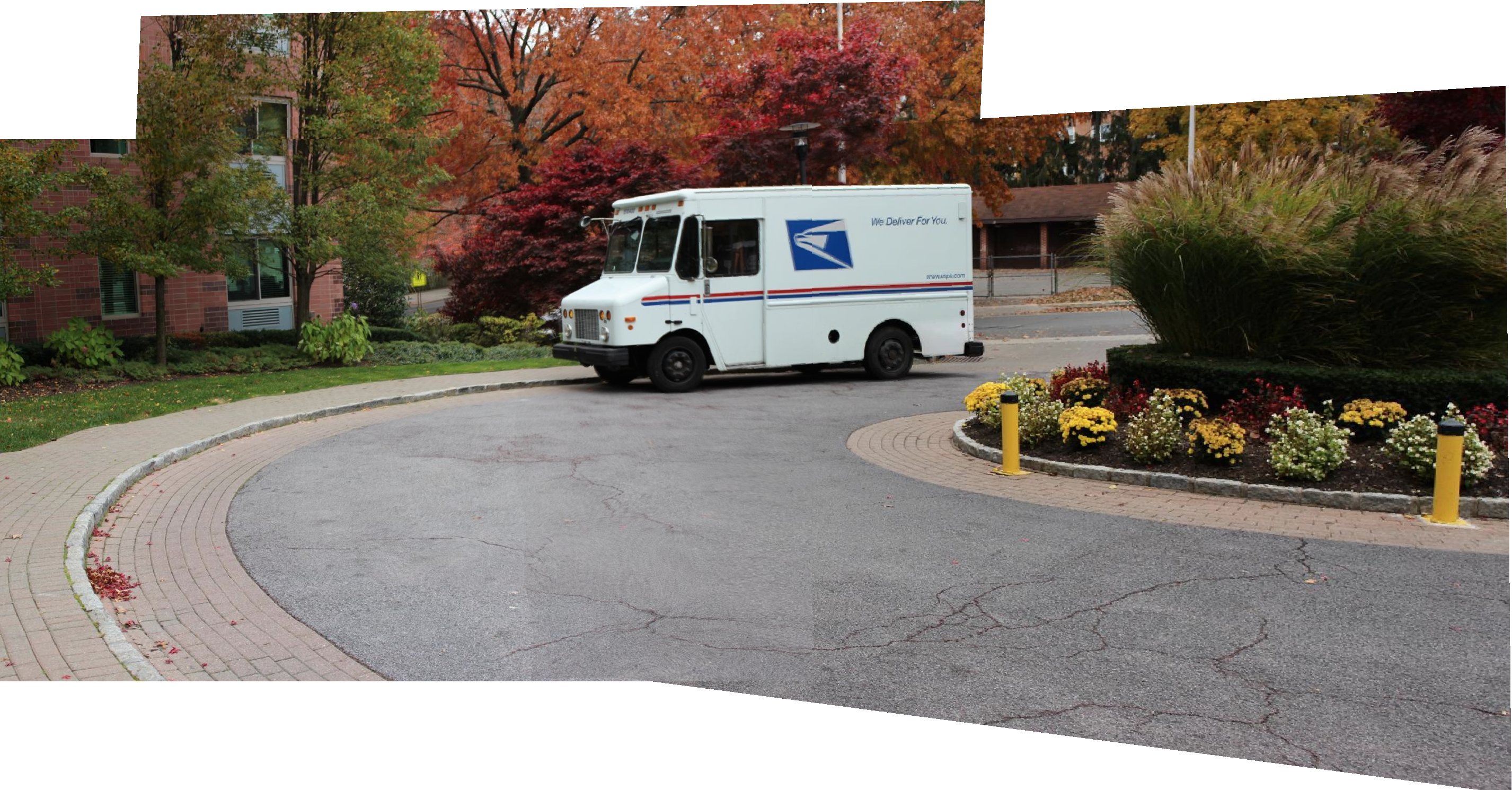}};
        \spy [closeup_w,magnification=2.5] on ($(FigA)+(-0.10,0.135)$) 
        in node[smallwindow,anchor=south] at ($(FigA.south) + (-0.10,0.135)$);
        \spy [closeup_r,magnification=2.5] on ($(FigA)+(-0.47,-0.13)$)
        in node[smallwindow,anchor=south] at ($(FigA.south) + (-0.47,0.32)$);
        \spy [closeup_r,magnification=1.5] on ($(FigA)+(0.08,0.180)$)
        in node[smallwindow,anchor=south] at ($(FigA.south) + (0.40,0.115)$);
        \end{tikzpicture}
    }{REwarp (ours)}
    \vspace{-5pt}
    \caption{\textbf{Comparison on Multiple image stitching.}}
    \label{fig:multiview}
    \vspace{-8pt}
\end{figure}

\noindent\textbf{Multiple image stitching}
Sequential inferences with REwarp enable multiple image stitching. In \cref{fig:multiview}, we demonstrate a qualitative comparison of multi-view stitching between LPC \cite{jia2021leveraging} and \textit{ours}. To evaluate the image, we determine a reference image $\mathbf{I}_r$ and two target images $\mathbf{I}_{t_1}, \mathbf{I}_{t_2}$. Then we sequentially estimate transformations between three frames $\mathbf{I}_r$, $\mathbf{I}_{t_1}$ and $\mathbf{I}_r$, $\mathbf{I}_{t_2}$. Despite the relaxed discontinuity of our constraint, REwarp may be exposed to the discontinuity as in the upper crop of trucks in each figure, slightly.

\begin{figure}[t]
    \begin{tikzpicture}[x=6cm, y=6cm, spy using outlines={every spy on node/.append style={smallwindow_w}}]
        \node[anchor=center] (FigA) at (0,0) {\includegraphics[width=0.95\linewidth, height=0.60\linewidth]{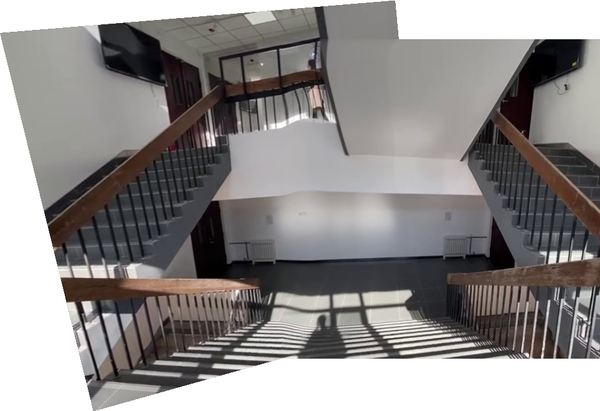}};
        \draw[red, line width=0.50mm] (-0.04,-0.20) rectangle (0.10,0.35);
    \end{tikzpicture}
    \vspace{-15pt}
    \caption{\textbf{Failure Case.}}
    \label{fig:failure_case}
    \vspace{-15pt}
\end{figure}

\noindent\textbf{Under Large Parallax}
In \cref{fig:failure_case}, we explore REwarp's stitching with images exposed to large parallax errors. As shown in the figure, it mostly guarantees continuity while providing visually unnatural image composition. Because TPS is a global transformation with smooth warps, our constraint may cause unnecessary warps in other local regions. To overcome this, a more advanced TPS warps with higher flexibility and C2 continuity in boundary would be required.

\section{Conclusion}
We proposed a Recurrent Elastic warp (REwarp) for parallax-tolerant deep image stitching, which recursively and sequentially estimates global and residual warps for misalign correction. REwarp resolves the hole generation and discontinuity issues of the previous work that leverages deep elastic warps and successfully demonstrated its image stitching on traditional real images and UDIS-D dataset quantitatively achieving 1.36dB mPSNR gain compared to Robust ELA. Furthermore, REwarp's fast computation and perfect success ratio in our experiment validated the potential usage for real-time image stitching.


{\small
\bibliographystyle{ieee_fullname}
\bibliography{egbib}
}

\end{document}